# Design of a PCIe Interface Card Control Software Based on WDF

Meng Shengwei
Automatic test and control institute
Harbin institute of technology
Harbin, China
mengsw@hit.edu.cn

Lu Jianjie
Automatic test and control institute
Harbin institute of technology
Harbin, China
jacklu@hit.edu.cn

*Abstract*—Based on the clear analysis of the latest Windows driver framework-WDF, this paper has implemented a driver of the PCIe-SpaceWire interface card device and put forwards a discussion about ensuring the stability of PCIe driver. At the same time, Qt and OpenGL are used to design the upper application. Finally, a functional verification of the control software is provided.

*Keywords*—**WDF, driver, PCIe, Qt**

## I. INTRODUCTION

The PCIe-SpaceWire interface card is a standard PCIe device and can be installed on a desktop computer, supporting a SpaceWire network system for controlling SpaceWire network communications. A study by Stanford University shows that device drivers have error rates up to three to seven times higher than the rest of the kernels [1].So it is critical that developers can implement a hardware driver correctly and efficiently in a control software.

In order to render the full performance of the superiority of the PCIe and to ensure the stability of the control software, a driver was implemented by using WDF. The GUI application was developed by using Qt and OpenGL. The whole control software can achieve control functions including link enable, link reset, port found and data acquisition. In addition, as a PCIe control software, a PCIe control panel is provided to read and write data directly and a 3D model is designed to display the three axis accelerometer according the data in the PCIe card.

## II. WDF OVERVIEW

For many years, drivers for Windows operating system should be implemented with the Windows Driver Model (WDM). However, the steep learning curve of WDM has limited driver development. And errors in drivers often cause some system crash [2].

WDF provides a new driver development method by repackaging WDM. Developers are able to focus on the interaction of hardware rather than the complexities of the operating system. WDF provides an abstraction layer between the driver and operation system where the driver interact with the framework for most services. So WDF greatly improves the stability of the system. A conceptual view of the WDF architecture is shown in figure 1.

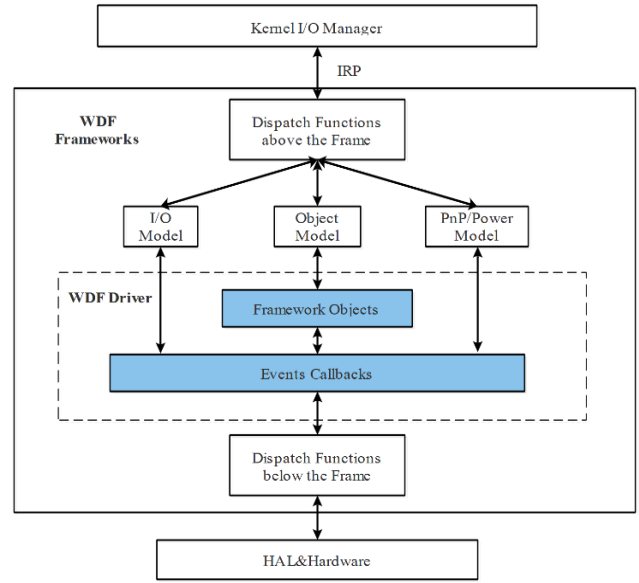

Figure 1. A Conceptual View of the WDF Architecture [3]

User defined routines and objects need to be registered in the framework. The framework will complete almost all of the remaining tasks, including managing object lifetimes, PnP&Power state and the flow of I/O requests, tracking objects that are in use and freeing them when they are no longer required.

WDF consists of two frameworks-UMDF (User Mode Driver Framework) and KMDF (Kernel Mode Driver Framework). In order to perform some kernel operation, KMDF is used to design the driver

## III. PCI EXPRESS DRIVER DESIGN

The design of drivers is in event driven based on WDF. So the flow of the driver is designed according to triggered events, the equipment insertion and the equipment discovery, as shown in Figure 2

When the PCIe driver is first loaded by Windows operating system, the DriverEntry routine, which is the entry of the driver, will be called. After the detection of the new PCIe-SpaceWire interface card into the desktop computer by PNP manager, WDF framework calls Spw_PCIeEvtDeviceAdd callback routine to complete the initialization of the driver, including creating device objects, getting system resources, creating GUID interface, setting I/O transmission methods, initializing I/O queue and registering PnP&Power Routines.

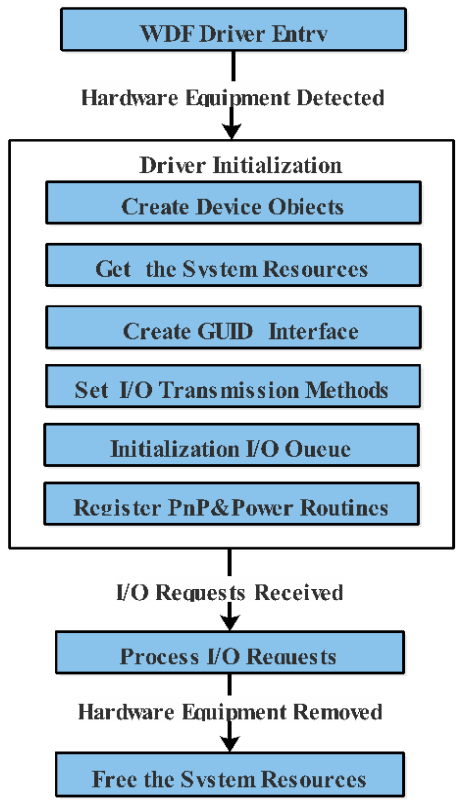

Figure 2. Driver Design Flow

After Windows finds the PCIe interface card, WDF calls the Spw_PCIeEvtDevicePrepareHardware routine to get the resources. Conversion from physical address to virtual address should be finished in MmMapIoSpace function.

WDF provides three I/O queues: Read, Write, and IoControl. In this paper, in order to reduce the difficulty of design, all the read and write requests are sent in the form of IoControl queue. Spw_PCIeEvtIoDeviceControl is the routine to process I/O request. It performs different tasks according to the control words from the application, including getting the BAR0 physical starting address, reading and writing registers.

When the card is removed, WDF will automatically release the memory space of the device by calling the Spw_PCIeEvtDeviceReleaseHardware routine. Drivers must release the allocated memory space, otherwise it may cause the memory overflow or even the operating system crash. As for the PnP&Power management, WDF provides intelligent defaults for common operations.

The INF file provides important information for the installation and identification of driver. In this paper, INF file marked the driver version and author, and also defined the Windows standard multifunction adapter as the device class. Some information in device manager is shown in the figure 3 and figure 4.

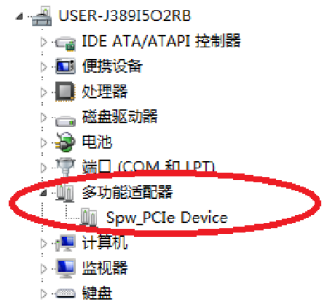

Figure 3. Defined as the Windows Standard Multifunction Adapter Class

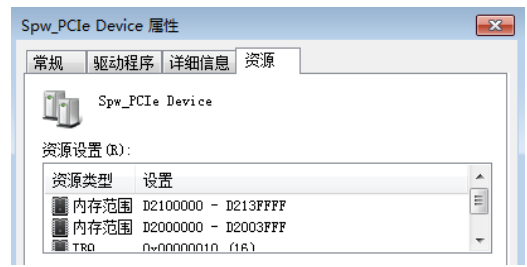

Figure 4. BAR0 and BAR2 Physical Address Range

Here we gives some tips to ensure the stability of WDF drivers. Modern operating systems, such as Linux, Windows use paging mechanism in memory management. The paged memory can be exchanged to the hard disk, while the non-paged memory cannot. Developers must ensure that the memory page of the program code interrupt request priority higher than DISPATCH_LEVEL (including DISPATCH_LEVEL) [4] is on the non-paged memory, otherwise it will cause the system to hang up. So in WDF, developers must pay more attention when using macro PAGE_CODE to tag a routine on the paged memory. Meanwhile, developers also should be careful that reading and writing memory cannot be crossed.

## IV. APPLICATION DESIGN

There are two important concepts when designing the communication between applications and drivers, that is, GUID (Globally Unique Identifier) and CTL_CODE macro.

GUID is a global unique identifier to generate a set of 128 binary number to identify a certain entity by using a specific algorithm (such as the time or place, etc.). The application can find its corresponding driver according to the GUID value. CTL_CODE is a macro for creating a unique 32 bit system I/O code, which is the second parameter of the I/O processing routine DeviceIoControl and includes 4 parts: device type (device type, high 16 bit (16-31)), access (access restrictions, 14-15), function (function 2-13) method (I/O access memory usage). Figure 5 shows the design flow of the application.

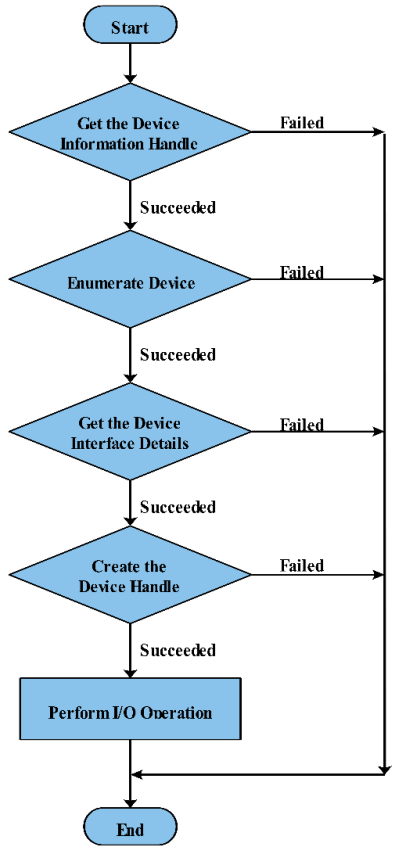

Figure 5.  Design Scheme of the Module

Graphical interface programs should be easy to use, so that it can restrict user's false operation effectively [5]. In this paper, the graphical main panel designed by using Qt is shown in figure 6.

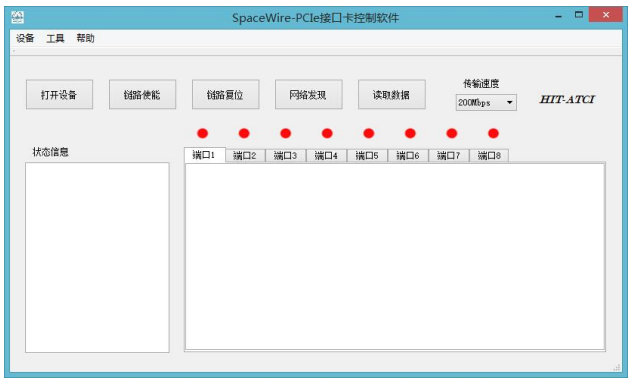

Figure 6.  The Graphical Main Panel

Considering the importance of software stability, the application is designed in multi-process control method. Each control command will execute an independent process. By the way, it is recommended to use encapsulated class QProcess in Qt. Using its member function can call an external program easily.

## V. VERIFICATION

A verification platform is consisted of the SpaceWire three axis accelerometer, the standard SpaceWire router and the PCIe-SpaceWire interface card. Figure 7 and figure 8 show the structure and physical map.

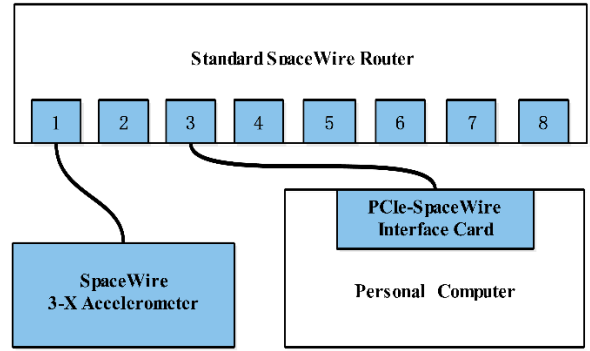

Figure 7.  Verification Network Structure

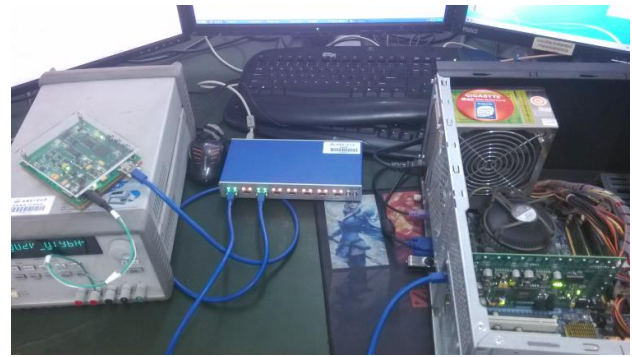

Figure 8.  Physical Map

First, the physical starting address of the interface card BAR0 0xD2100000 is acquired, which is same as the value in the device manager shown as figure 9. Data in the memory with 100 offset address is 0 and then write data 2222 to it. As figure 9 and figure 10 shows, the read data become 2222. So it is verified that driver can achieve reading and writing the PCIe device.

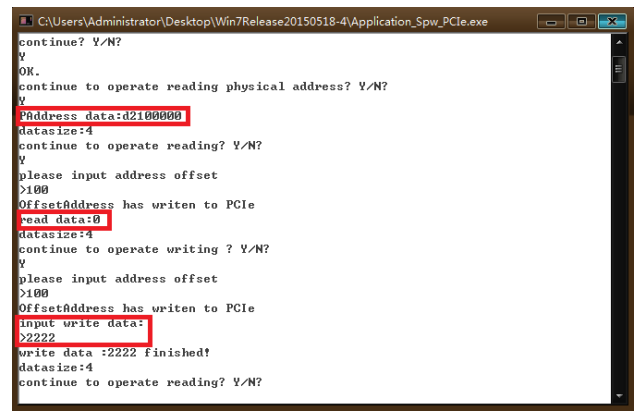

Figure 9.  PCIe Read and Write Functional Verification Ⅰ

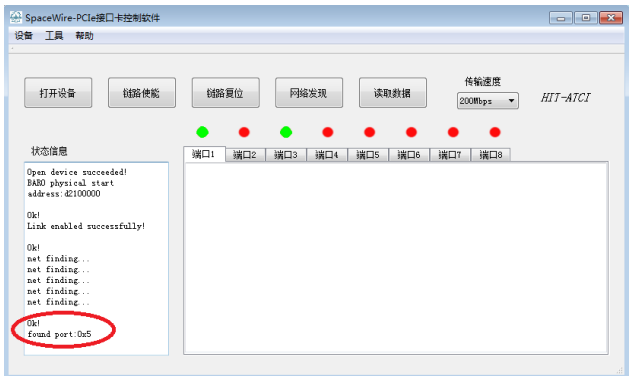

Figure 10. PCIe Read and Write Functional Verification Ⅱ

After finding ports, we get 0x05(101B) as shown in figure 11, which means the first and third ports are used.

Figure 11. Port Discovery

The rotation speed and direction of the three axis accelerometer 3D model is associated with the received data in PCIe device as shown in figure 12 and 13, where the model is provided by Qt examples.

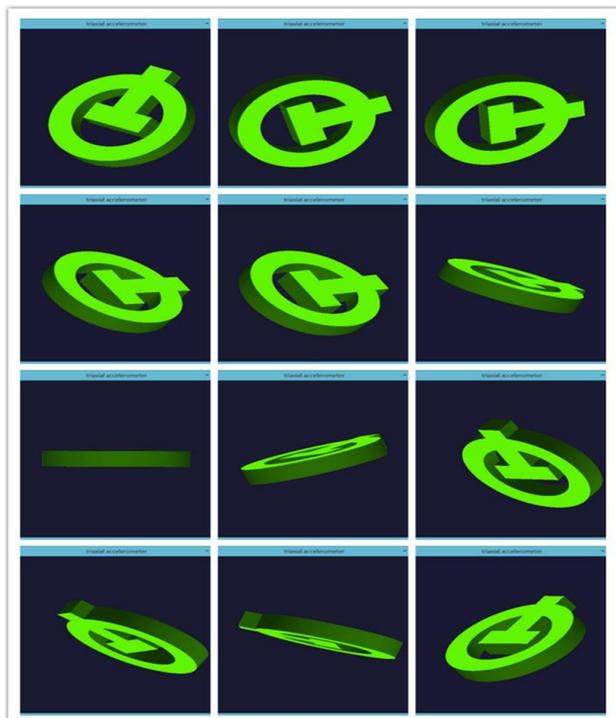

Figure 12. The Different Rotation Direction of the 3D Model according to the Data in the PCIe Card.

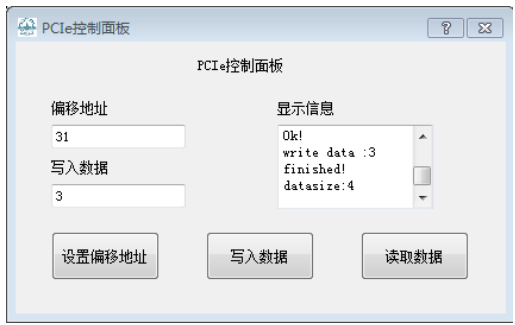

Figure 13. Write Different Data in the PCIe Card

## VI. CONCLUTION

This paper provides a detailed summary of WDF and gives a design of the PCIe device control software based on WDF. The design having been verified is guiding significance to achieve a stable and efficient PCIe control software.